\documentclass[twoside,11pt]{article}

\usepackage{jmlr2e}
\usepackage{url}
\usepackage{lastpage}
\usepackage{needspace}

\newcommand{\tuiml}{\textsc{TuiML}}

\ShortHeadings{\tuiml: Machine Learning for AI Agents}
  {Verma, Lim, Bifet, and Pfahringer}
\firstpageno{1}

\begin{document}

\title{TuiML: Machine Learning for AI Agents}

\author{\name Nilesh Verma$^1$ \email nilesh.verma@waikato.ac.nz \\
       \name Nick Lim$^1$ \email nick.lim@waikato.ac.nz \\
       \name Albert Bifet$^{1,2}$ \email albert.bifet@waikato.ac.nz \\
       \name Bernhard Pfahringer$^1$ \email bernhard.pfahringer@waikato.ac.nz \\
       \addr $^1$AI Institute, University of Waikato\\
       Hamilton 3216, New Zealand\\
       $^2$LTCI, T\'el\'ecom Paris, Institut Polytechnique de Paris\\
       19 place Marguerite Perey, 91120 Palaiseau, France}

\editor{N/A}%TODO Editor Name}
\maketitle

\begin{abstract}%
Machine-learning libraries such as Weka and scikit-learn were designed for
human programmers. Language-model agents now use these same libraries by
recalling APIs from memory and writing code, an approach that hides what a
library offers, delays errors until runtime, and loses experimental state
between turns. We present \tuiml, a self-contained machine-learning
library built for AI agents, with native algorithms across supervised,
unsupervised, time-series, data handling, tuning, and
evaluation tasks. Every component describes itself through machine-readable
metadata and parameter schemas, so an agent can search the library,
inspect components, compose validated workflows, and register new ones
that become discoverable in turn. Every call is validated, seeded, and
traced, and sessions export as runnable notebooks, making experiments
reproducible by construction. One specification layer drives the Model
Context Protocol (MCP), agent-framework adapters, a \texttt{Python} API, a
CLI, and local model serving, while data and models never leave the
machine. Benchmarks show \tuiml{} remains predictively competitive with
scikit-learn and Weka. 
%It stays a conventional library for humans, yet is
%designed first for agents that read, extend, and operate machine learning
%on their own.
While looking like a conventional library to a human user, \tuiml{} is designed for agents first, allowing them to read, extend, and operate machine learning autonomously.
\tuiml{} is open source, with documentation at
\url{https://tuiml.ai}.
\end{abstract}


\section{Introduction}
\label{sec:intro}
Machine-learning libraries have always been written with a particular user in
mind, namely a programmer who reads the documentation, learns the API, and
assembles scripts by hand. Weka, scikit-learn, and their successors all share
this assumption. However, language-model agents operating these same
libraries break this assumption, and the failures are measurable. The
strongest system reported in MLE-bench reaches a Kaggle bronze-medal
threshold on only 16.9\% of competitions \citep{chan2025mlebench},
MLAgentBench identifies hallucination and long-horizon planning as persistent
failure modes \citep{huang2024mlagentbench}, and agents asked to recall large
interfaces invent plausible but non-existent calls, a problem that motivated
retrieval over live documentation \citep{patil2024gorilla}. The mismatch is
structural, since agents are expected to hold an interface in memory that was
never designed to be held that way.

We present \tuiml, a comprehensive machine-learning library designed from the
ground up for AI agents, addressing this mismatch at the level of the library
rather than the agent. Instead of asking an agent to generate an entire
program, \tuiml{} exposes machine-learning operations as typed, discoverable
actions, guided by the simple principle that an agent need not memorize a
library when the library describes itself. Every algorithm is a
schema-described action an agent can call directly. A queryable registry
makes components discoverable by task, data shape, or constraint. A
validated execution layer with structured errors, recorded seeds, and
replayable notebook cells makes agent-driven experiments trustworthy and
reproducible by construction. The Model Context Protocol (MCP) standardizes
tool discovery and schema-based invocation \citep{anthropic2024mcp}, and
\tuiml{} supplies the machine-learning semantics, persistent state, and local
runtime behind those calls. The design builds on the interface discipline of
scikit-learn \citep{pedregosa2011scikit,buitinck2013api}, Weka
\citep{hall2009weka}, MOA \citep{bifet2010moa}, and River
\citep{montiel2021river}. It supports run-time discovery, machine-readable
parameter schemas, and task-level execution. Unlike approaches that
consolidate agent actions into executable code \citep{wang2024codeact},
\tuiml{} constrains actions to validated calls, eliminating silent argument
errors while retaining a \texttt{Python} interface through which a
code-acting agent can import the components.

\section{Design and Architecture}
\label{sec:design}
\begin{figure}[t]
  \centering
  \includegraphics[width=\linewidth]{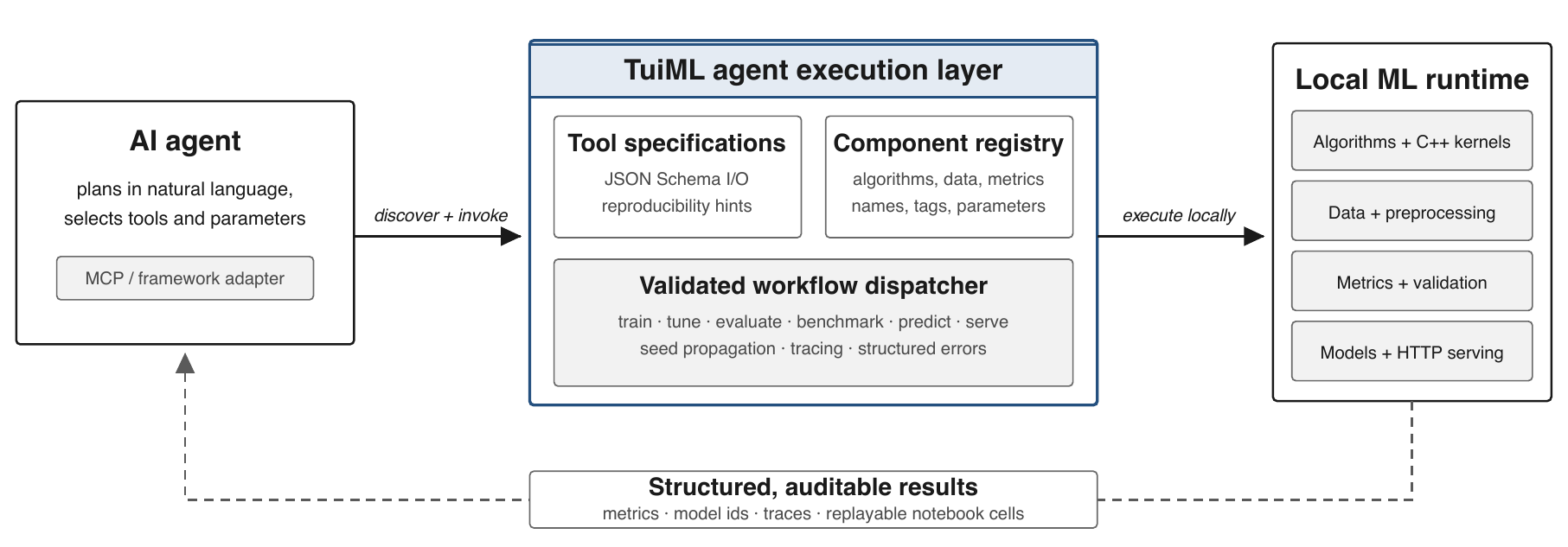}
  \caption{\tuiml{} separates agent planning from machine-learning execution
  through one typed contract, while data and fitted models remain local. Solid
  arrows show the forward path, where the agent discovers and invokes MCP tools
  that the execution layer validates and runs locally. Dashed arrows show the
  return path, where metrics, model identifiers, traces, and replayable
  notebook cells flow back to the agent.}
  \label{fig:architecture}
\end{figure}
Every component in \tuiml{} is addressed by a registry name and a dictionary
of constructor parameters, a single scheme shared by algorithms,
transformations, and metrics. The registry resolves each name to an
implementation and exposes its metadata and a JSON Schema for its parameters,
extending the uniform estimator protocol of conventional libraries
\citep{buitinck2013api} with run-time discovery. Because discovery and
execution consult the same registry, a newly registered component becomes
searchable and agent-callable immediately, and invalid names, misplaced
parameters, and unsupported estimators are rejected with structured errors
rather than silently substituted. As shown in
Figure~\ref{fig:architecture}, an agent connects through an MCP client,
discovers the available \tuiml{} tools with their schemas, and submits
workflow specifications covering the full experimental cycle, from data
inspection and preprocessing through training, tuning, evaluation, serving,
and notebook export. The agent decides what to try, and \tuiml{}
decides how the experiment is instantiated and recorded.

\begin{figure}[t]
  \centering
  \includegraphics[width=\linewidth]{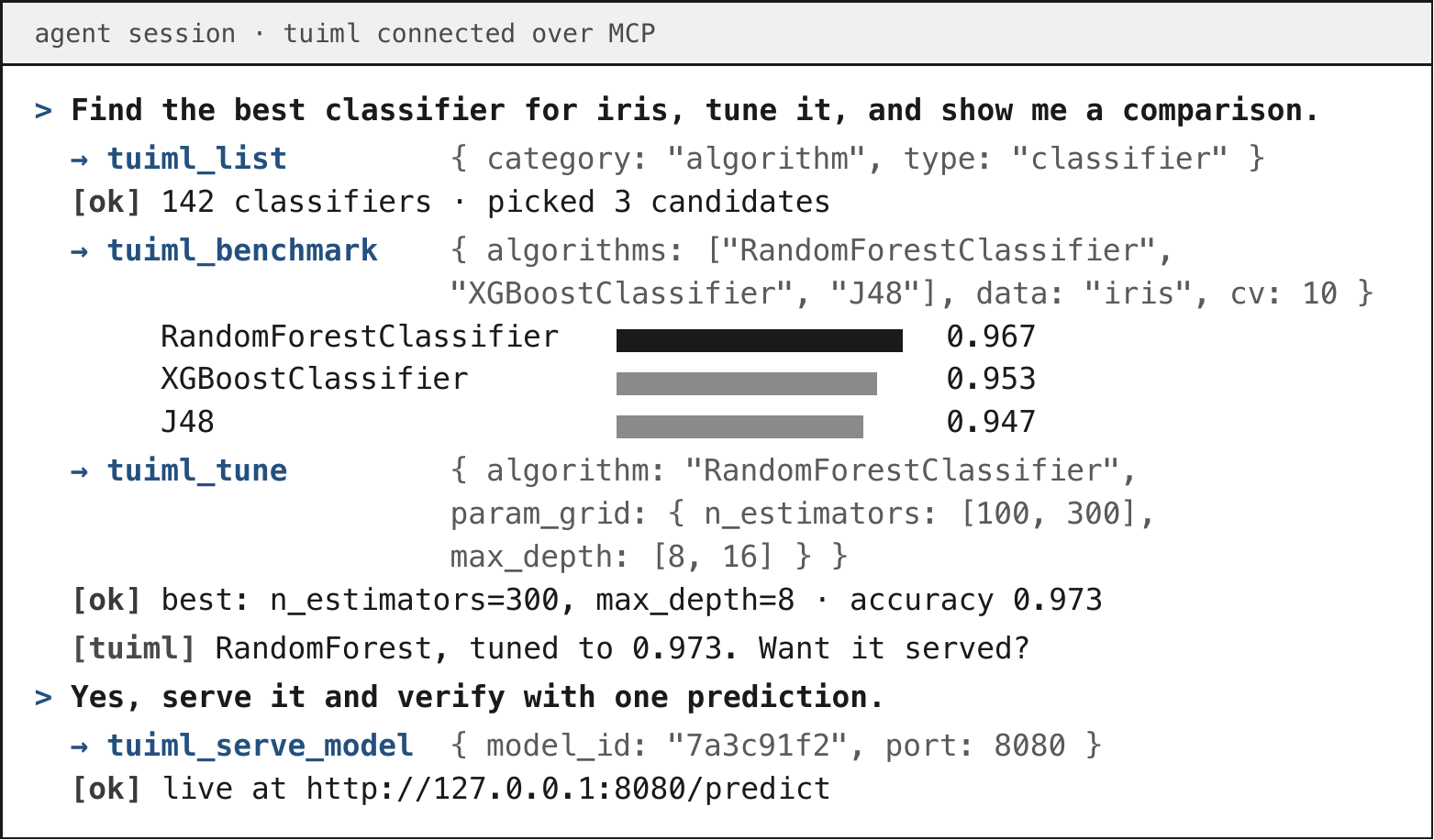}
  \caption{An agent session against \tuiml. One request resolves into typed
  calls that discover, benchmark, tune, and serve a model, with \tuiml{}
  validating, seeding, and recording each step.}
  \label{fig:session}
\end{figure}

Figure~\ref{fig:session} traces a representative session, in which the agent
queries the registry for candidate classifiers, benchmarks them under
cross-validation, tunes the winner, and serves the fitted model, with every
argument validated before any model is fitted. Reproducibility follows from
the same contract. \tuiml{} records each successful call with its seed and
model lineage and exports the sequence as executable notebook cells, so every
session yields a runnable artifact by construction
\citep{pineau2021reproducibility}. Compact JSON Lines traces retain
timestamps, arguments, duration, and status without copying large arrays.
Datasets and fitted models never leave the machine, with the agent receiving
schemas and structured summaries rather than raw data.

\section{Implementation and Evaluation}
\label{sec:implementation}

\begin{figure}[t]
  \centering
  \includegraphics[width=\linewidth]{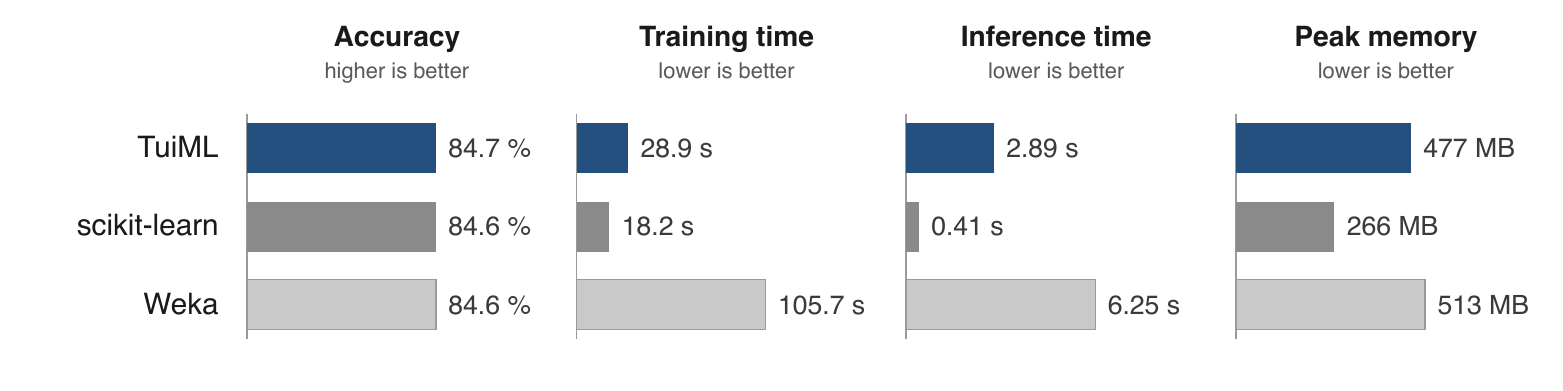}
  \caption{Matched results on 51 TabArena datasets, averaged over 3,318 runs.
  Accuracy covers the 2,587 classification runs, and Weka memory includes its
  JVM baseline.}
  \label{fig:benchmark}
\end{figure}

\tuiml{} is a \texttt{Python}~3.10+ library built on \texttt{NumPy}
\citep{harris2020array}, with performance-critical kernels compiled from
\texttt{C++} through \texttt{pybind11}. It provides classification,
regression, clustering, association mining, anomaly detection, and
time-series modeling, together with preprocessing, feature engineering,
metrics, statistical tests, tuning, and reporting. Optional scikit-learn
\citep{pedregosa2011scikit} and CapyMOA \citep{gomes2025capymoa} learners are
exposed through namespaced wrappers. Unlike AutoML systems that prescribe
a search policy \citep{feurer2015autosklearn,trirat2024automlagent}, \tuiml{}
supplies a stable action space on which such systems can be built. To evaluate the runtime, we compare the thirteen algorithms shared by
\tuiml, scikit-learn, and Weka on 51 TabArena v0.1 datasets
\citep{erickson2025tabarena} hosted on OpenML \citep{vanschoren2014openml},
with hyperparameters and ten-fold splits aligned across frameworks and runs
executed in isolated, single-threaded processes. As shown in
Figure~\ref{fig:benchmark}, \tuiml{} achieves accuracy comparable to
scikit-learn and Weka while being competitive in runtime and
memory consumption, thus demonstrating that an agent-facing interface orchestrate a competitive runtime.

\section{Conclusion}
\label{sec:conclusion}

\tuiml{} provides a self-describing machine-learning library in which
components are discoverable, calls are validated, and experiments are
traceable, backed by native, performance-oriented implementations across
classification, regression, clustering, anomaly detection, association
mining, and time-series modeling. Documentation, tutorials, contribution
guides, and the full benchmark protocol are available at
\url{https://tuiml.ai}. The library ships with a comprehensive test suite
and cross-platform builds for Linux, macOS, and Windows. \tuiml{} remains
alpha software, and a syntactically valid call can still encode a
statistically inappropriate experiment, so results warrant the usual
scrutiny. Future work includes a controlled comparison of schema-guided calls
against free-form code under matched models and budgets, broader algorithm
coverage, and measurement of tool-selection accuracy and token cost. We
expect \tuiml{} to evolve as a comprehensive tool for research and real-world
applications as agent interfaces mature.

% Acknowledgements and the funding / competing-interests declaration are
% required by JMLR and must appear before any appendices and the references.
\needspace{5\baselineskip}
\acks{\tuiml{} was developed at Te Ipu o te Mahara, the Artificial Intelligence
Institute of the University of Waikato, and builds on the tradition of open
machine-learning software established there by Weka. This work was supported in
part by the TAIAO programme (Time-Evolving Data Science and Artificial
Intelligence for Advanced Open Environmental Science), funded by the New Zealand
Ministry of Business, Innovation and Employment.}

\vskip 0.2in
\bibliography{refs}

\end{document}